\documentclass[conference]{IEEEtran}
\IEEEoverridecommandlockouts
\usepackage{cite}
\usepackage{amsmath,amssymb,amsfonts}
\usepackage{algorithmic}
\usepackage{graphicx}
\usepackage{amsthm}
\usepackage{censor}
\usepackage{hyperref}
\usepackage{textcomp}
\usepackage{tabularx}
\usepackage{titlesec}
\usepackage{threeparttable}
\usepackage{xcolor}
\usepackage{float}
\usepackage{booktabs}
\usepackage{arydshln}

\begin{document}



\title{Can Low-Rank Knowledge Distillation in LLMs be Useful for Microelectronic Reasoning?   \\
}
\author{
\IEEEauthorblockN{Nirjhor Rouf$^*$}
ECE Department, \\ North Carolina State University. \\
nrouf2@ncsu.edu
\and
\IEEEauthorblockN{Fin Amin$^*$}
ECE Department, \\ North Carolina State University. \\
samin2@ncsu.edu
\and
\IEEEauthorblockN{Paul D. Franzon}
ECE Department, \\ North Carolina State University. \\
paulf@ncsu.edu
}


\maketitle

\def\thefootnote{*}\footnotetext{These authors contributed equally to this work.}

\footnotetext{To appear in IEEE International Workshop on LLM-Aided Design (LAD'24) }
\renewcommand{\thefootnote}{\arabic{footnote}}
\setcounter{footnote}{0}

\begin{abstract}
In this work, we present empirical results regarding the feasibility of using offline large language models (LLMs) in the context of electronic design automation (EDA). The goal is to investigate and evaluate a contemporary language model’s (Llama-2-7B) ability to function as a microelectronic Q\&A expert as well as its reasoning, and generation capabilities in solving microelectronic-related problems. Llama-2-7B was tested across a variety of adaptation methods, including introducing a novel low-rank knowledge distillation (LoRA-KD) scheme. Our experiments produce both qualitative and quantitative results. Furthermore, we release our evaluation benchmark along with the code necessary to replicate our experiments at \href{https://github.com/FinAminToastCrunch}{github.com/FinAminToastCrunch}. 
\end{abstract}


\section{Introduction and Motivation}

The emergence of Large Language Models (LLM) has revolutionized the field of natural language processing. At present, LLMs are garnering significant research interests for domain-specific tasks. In the field of electronic design automation (EDA) in particular, applications of LLMs are still at the nascent stage. However, it is very apparent that the effective use of LLMs in EDA can improve manufacturing yields by streamlining the design flow when it comes to IC design. Recently published works showed the successful use of LLMs in chip design \cite{ChatEDA,ChipNeMo,ChipChat}. Additionally, LLMs have also shown significant proficiency in the analysis of designed systems \cite{trojan} and even in reviewing and analysis of design specifications of VLSI systems \cite{SpecLLM}. Development of open-source benchmarks such as VerilogEval \cite{VerilogEval} is also facilitating future research in this field. Similarly, LLMs can be useful in enhancing productivity. Internal studies carried out at Nvidia have shown that checklist related tasks can take up to 60\% of an engineer's time and thus bottleneck productivity \cite{ChipNeMo}. An LLM-based engineering assistant can certainly reduce this bottleneck by helping with engineering knowledge dissemination.

\newlength{\origabovecaptionskip}
\newlength{\origbelowcaptionskip}
\begin{figure}[t] 
  \setlength{\origabovecaptionskip}{\abovecaptionskip}
  \setlength{\origbelowcaptionskip}{\belowcaptionskip}
  \setlength{\abovecaptionskip}{0pt}
  \setlength{\belowcaptionskip}{12pt}
  \centering
  \includegraphics[trim=1cm 4cm 3cm 3cm, clip=true, width=\linewidth]{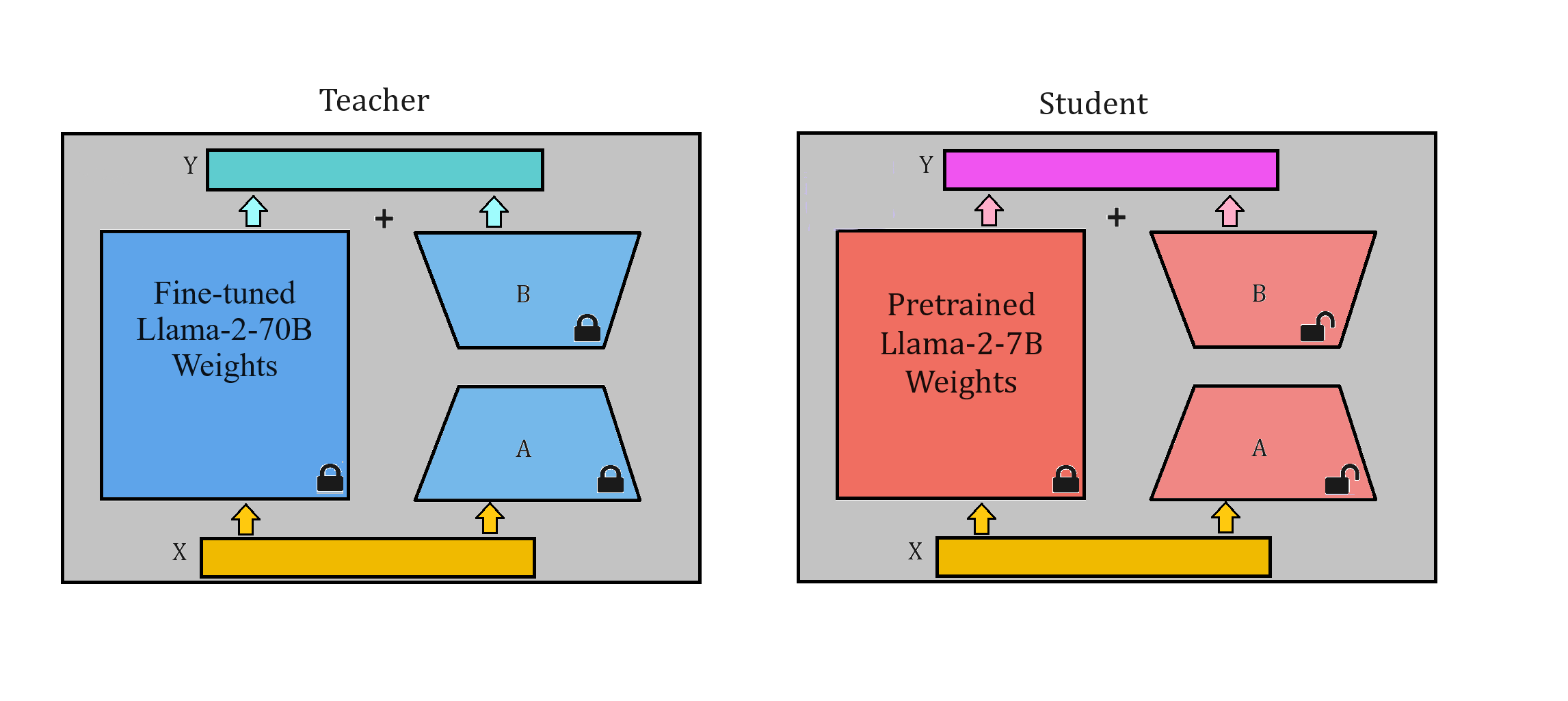}
  \caption{LoRA-KD works by first fine-tuning the teacher model using LoRA. Afterward, the teacher is frozen and its outputs are used for equation \ref{eq: kd_loss}. Note that only the low-rank $A$ and $B$ parameters of the student are updated.}
 
  \label{fig:lora_kd}
  
\end{figure}

However, several key challenges must be addressed for more effective and efficient application of LLMs in EDA. One big concern is the unintentional data retention of DNNs from training sets \cite{KDinAssessment}. There are two aspects of this issue. Firstly, classified IP designs can be leaked if the API stores user input. Secondly, when trying to complete a user request, the LLM can inadvertently use copy-righted IP designs without attributing references to them--potentially causing downstream legal trouble. Another major challenge is the heavy computational resource requirement of LLMs. For example, Meta's Llama2-70B requires 130 GB memory to load \cite{llama2}.  

\begin{figure*}[th] 
  \centering
  \includegraphics[width=.95\linewidth]{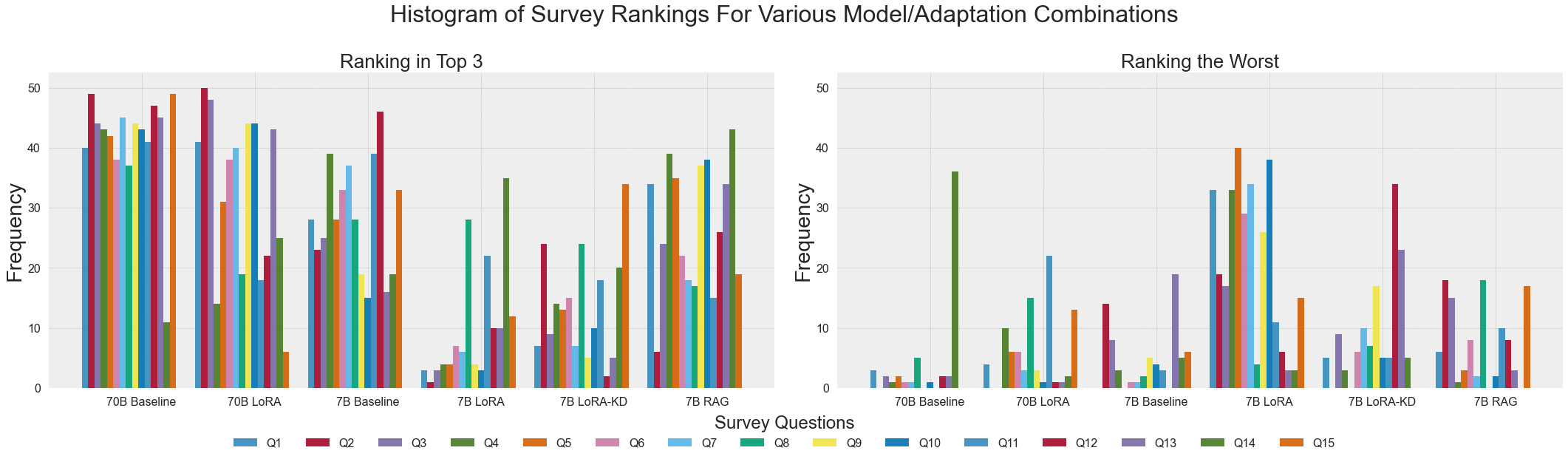}
  \caption{These charts show histograms of which configurations were ranked in the top half and declared the worst according to third-year microelectronics students. Survey participants had to order the outputs of each configuration on 15 questions. A total of 51 rankings were considered after filtering for quality.}
  \label{fig:survey}
\end{figure*}

Choosing the appropriate LLM for EDA applications is also a big challenge and here, the proprietary vs open-source debate must be addressed. While proprietary models, such as ChatGPT-4 \cite{gpt4} are powerful, they have limited accessibility, store user data/designs, and are pay-to-use. Additionally, the inability to fine-tune them hinders their capabilities in domain-specific EDA tasks. On the other hand, open-source LLMs offering better accessibility are restricted by limited scale and resources compared to their proprietary counterparts resulting in lower performance \cite{KDsurvey}. In this work, we explore the feasibility of adapting the open-source Llama-2-7B for use in EDA education. We focus on this model in particular because it can be used on consumer hardware. Our contributions are as follows:

\begin{enumerate}
    \item A quantitative and qualitative analysis of Llama-2-7B adapted in various ways for EDA usage. This investigation allows us to understand the impact of fine-tuning, distillation, and retrieval augmentation on the model's performance in the context of EDA knowledge.
    
    \item  We introduce and evaluate a novel fine-tuning method, Low-Rank Knowledge Distillation (LoRA-KD).
    
    \item The release of a benchmark, \textit{RAQ}, designed for evaluating LLMs on EDA knowledge, aimed at facilitating future research and development in the field. 
\end{enumerate}

\section{Prior Work on LLMs for EDA}

 In the burgeoning field of EDA, early explorations into the applications of LLMs have already returned promising results, particularly in the nuanced areas of chip design, debugging, and script generation. Recently developed LLM-powered ChatEDA is capable of streamlining the IC design flow from RTL to GDSII \cite{ChatEDA}. ChatEDA integrates Automage, a fine-tuned LLM based on Llama-2-70B architecture, with an EDA tool. Automage serves as an interface that accepts human requests and manipulates the EDA tool through API for task completion. ChatEDA was tested on performance evaluation, parameter grid search, parameter tuning, customized optimization and clock period minimization. 

On the other hand, Nvidia took a slightly different approach with their ChatNeMo, a Llama-2 based LLM for chip design which contributes greatly to improving productivity as an engineering chatbot assistant. It is also capable of generating EDA scripts and, bug summarization and analysis \cite{ChipNeMo}. ChatNeMo outperforms GPT-4 at engineering assistant chatbot and EDA script generation tasks while showing comparable performance at bug summarization and analysis whereas ChatEDA has shown comparable or better performance than GPT-4 in all its evaluated cases.

Another work \cite{ChipChat} explores the possibility of LLM applications in conversational hardware design by having a hardware engineer co-architect a microprocessor architecture with GPT-4 and this design was sent to tapeout. In addition to these, the possibility of LLM applications in generating VLSI design specifications has also been explored. SpecLLM has shown significant proficiency in assisting engineers in generating and reviewing architecture specifications \cite{SpecLLM}.

Hardware security assessment is one more field which has studied the feasibility of language models. The authors of \cite{trojan} present an automated flow to identify suitable modules in large HDL databases for hardware trojan insertion using a general-purpose LLM. The model's ability to pinpoint candidate modules for the attack can be indicative of its significant comprehension of RTL codes and system design.

\begin{table*}[ht]
\centering
\caption{Configurations' Performance on Reasoning and Accuracy Questions}
\begin{tabular}{l|lllllll}
\hline 
\textbf{RAQ: Reasoning} & \textbf{Ground Truth} & \textbf{70B Baseline} & \textbf{70B LoRA} & \textbf{7B Baseline} & \textbf{7B LoRA} & \textbf{7B LoRA-KD} & \textbf{7B RAG} \\
\hline
1a & Increase & Increase & Increase & Increase & Increase & Increase & Decrease \\
1b & 3 $\mu$m & 3 $\mu$m & 3 $\mu$m & 3 $\mu$m & 3 $\mu$m & 3 $\mu$m & 330 nm \\
2a & 0.775 mA & 5.58 mA & 5.58 mA & 1.28 A & 1.395 A & 0.06 A & \texttimes \\
2b & 1.55 V & 11.16 V & 8.6 V & 0.83 V & 0.647 V & 0.7 V & \texttimes \\
3a & 2 V & 2 V & 2 V & 5 V & 0.625 V & 5 V & 6 V \\
3b & 3 k$\Omega$ & 8 k$\Omega$ & 4 k$\Omega$ & 10 k$\Omega$ & 13 k$\Omega$ & 4 k$\Omega$ & \texttimes \\
4 & 2 & 2 & 2 & 2 & 2 & 2 & 2 \\
5a & 0.66 k$\Omega$ & 666.67 $\Omega$ & 2/3 k$\Omega$ & 0.5 k$\Omega$ & 3 k$\Omega$ & 0.5 k$\Omega$ & \texttimes \\
5b & 0.667 k$\Omega$ & 666.67 $\Omega$ & 0.5 k$\Omega$ & 0.25 k$\Omega$ & 1 k$\Omega$ & 1.6 k$\Omega$ & \texttimes \\
\hline
\hline
\textbf{RAQ: T/F Accuracy} & - & 84\% & 84\% & 72\% & 76\% & 76\% & 80\% \\
\hline 
\end{tabular}
\par
\vspace{5pt}
{\footnotesize Evaluated on the reasoning and T/F questions from the RAQ benchmark. Note that some of the reasoning questions required multi-step thinking, eg. \textit{based on your answer to part a, what is part b?} The $\times$ symbol denotes that the model refused to answer the question due to fallacious ``ethical reasons.''}
\label{table:reasoning_accuracy}
\end{table*}

\section{Adaptation Techniques for LLMs}

\subsection{Low-Rank Adaptation}

Low-rank adaptation (LoRA) addresses many issues associated with adapting LLMs for domain-specific usage\cite{Lora}. This method bypasses the expensive backpropagation of gradients across all parameters by keeping the backbone model frozen. This is done by assuming that the update to the model's weights have low-rank. In other words, instead of updating the backbone, we learn parameters $A$ and $B$ which learn the required changes to the output of the backbone. More explicitly, if we write the parameter update equation, LoRA makes the following approximation:      
\begin{align}
    \Theta_{t+1} &= \Theta_t - \eta \nabla_{\Theta} \mathcal{L}(\Theta_t) \\
     &\approx \Theta_0 - \alpha BA \\
     \text{i.e. } \eta \nabla_{\Theta} \mathcal{L}(\Theta_t) &\approx \alpha BA 
\end{align}
Where $\Theta \in \mathbb{R}^{d \times k}, B \in \mathbb{R}^{d \times r}, \text{ and } A \in \mathbb{R}^{r \times k}$. Note that $d \times k$ represents the size of the backbone model's (very large) parameter shapes. By selecting $r << min(d,k),$ LoRA provides a resource-efficient update to the backbone.

\subsection{Knowledge Distillation}

Knowledge Distillation (KD) \cite{knowledgedistillation} is a knowledge-transfer technique where a larger (teacher) network produces soft targets for a smaller (student) model. This can play a pivotal role in reducing the performance gap between larger and smaller models. Fine-tuning a smaller (student) model through KD can show improved performance compared to a normally fine-tuned small model. As an example, the authors of DistilBERT show that they can retain 97\% of the original BERT's performance despite a significantly smaller parameter count \cite{distilbert}. This indicates that a smaller model that can be deployed on weaker hardware, e.g. personal computer, can maintain feasibility in handling complex tasks related to EDA. Written explicitly, the loss used for KD is: 

\setlength{\abovedisplayskip}{0pt}
\begin{align}
    \mathcal{L}_{KD} &= (1-\alpha)\mathcal{L}_y(student(x), y) \nonumber \\ 
    &+ \alpha \mathcal{L}_{Dist}(student(x), teacher(x))  \label{eq: kd_loss}
\end{align}

Where $\mathcal{L}_{y}$ is the typical loss incurred between the student predictions and the target. $\mathcal{L}_{Dist}$ is the loss between what the student predicted and the teacher predicted on an input. More elaborate distillation techniques exist, for example, \textit{patient} KD aims at having the student mimic the teacher's intermediate layers in addition to the teacher's outputs \cite{Patient_kd}. We refer readers to \cite{KDsurvey} for further exploration.

\subsection{Low-Rank Knowledge Distillation (LoRA-KD)}

Although not entirely unprecedented, the combination of low-rank approximations and knowledge distillation is far less explored in the context of LLM fine-tuning. The authors of LoSparse \cite{losparse} introduce a new compression scheme for transformers \cite{attention} based on a truncated singular value decomposition. In their experiments, they find that combining this parameter compression scheme with knowledge distillation further improves performance. 

In our work, we reformulate this concept in accordance with figure \ref{fig:lora_kd}. We begin by fine-tuning the teacher (Llama-2-70B) using LoRA. Afterwards, we fine-tune the student (Llama-2-7B) via LoRA using $\mathcal{L}_{KD}$. We hypothesize that, if the updates to the teacher can be done in a low-rank fashion, then the underlying \textit{knowledge} being learned is also low-rank; therefore, the knowledge to be distilled to the student is also low-rank.

\begin{table*}[ht]
\centering
\caption{Comparison of Model/Adaptation Combinations Evaluated by Human Expert and GPT-4.5 Turbo via Likert Scale.}
\label{tab:model_performance_comparison}
\begin{tabular}{lcccccc}
\toprule
& \multicolumn{2}{c}{Human Expert} & \multicolumn{2}{c}{GPT-4.5 Turbo} & \multicolumn{2}{c}{Pearson Correlation} \\
\cmidrule(lr){2-3} \cmidrule(lr){4-5} \cmidrule(lr){6-7}
\textbf{Configuration} & \textbf{Accuracy} & \textbf{Quality} & \textbf{Accuracy} & \textbf{Quality} & \textbf{Accuracy} & \textbf{Quality} \\
\midrule
70B Baseline   & 4.2\textsubscript{$\pm$1.96} & 3.975\textsubscript{$\pm$1.96} & 4.625\textsubscript{$\pm$2.00} & 4.2\textsubscript{$\pm$1.95} & 0.51 & 0.43 \\
70B LoRA       & 4.35\textsubscript{$\pm$1.77} & 4.275\textsubscript{$\pm$1.84} & 5.7\textsubscript{$\pm$1.65} & 5.475\textsubscript{$\pm$1.67} & 0.47 & 0.51 \\
7B Baseline    & 3.4\textsubscript{$\pm$1.92} & 3.275\textsubscript{$\pm$1.87} & 4.15\textsubscript{$\pm$1.90} & 3.9\textsubscript{$\pm$1.93} & 0.53 & 0.57 \\
7B LoRA        & 3.5\textsubscript{$\pm$1.82} & 3.3\textsubscript{$\pm$1.73} & 4.2\textsubscript{$\pm$1.93} & 3.95\textsubscript{$\pm$1.90} & 0.66 & 0.73 \\
7B LoRA-KD & 3.525\textsubscript{$\pm$1.99} & 3.3\textsubscript{$\pm$1.83} & 4.475\textsubscript{$\pm$1.70} & 4.075\textsubscript{$\pm$1.60} & 0.60 & 0.59 \\
7B RAG         & 4.2\textsubscript{$\pm$1.87} & 3.7\textsubscript{$\pm$1.76} & 4.55\textsubscript{$\pm$1.96} & 4.15\textsubscript{$\pm$1.94} & 0.12 & 0.13 \\
\bottomrule
\end{tabular}
\par
\vspace{5pt}
\raggedright{ \footnotesize Each response to the 40 qualitative questions was evaluated on a 7-point Likert scale by a human expert and GPT-4.5 Turbo. $7$ denotes ``strongly agreed with'' and $1$ denotes ``strongly disagreed with.'' The subcolumns correspond to how much the evaluator agreed/disagreed with the accuracy/quality of the response. The standard deviations across the questions are written in sub-scripts. The correlation quantifies the consistency between the human expert and GPT-4.5 Turbo.}
\label{tab:likert}
\end{table*}

There are several advantages to doing this: 
\begin{itemize}
    \item As with ordinary RAG, a pre-trained model can be repurposed via hot-swapping the adaptation layer. For example, EDA educators can use LoRA-KD to learn separate (small) adaptation layers for English and Spanish in the context of a bilingual classroom.  
    \item KD has been used to enhance domain-adaptation tasks. We hypothesize that the \textit{dark knowledge} distilled from the teacher to the student will facilitate enhanced reasoning capabilities \cite{kd_dark_knowledge}.     \
    \item The training process remains fast. In our experiments, fine-tuning the student via LoRA-KD did not take much more time than ordinary LoRA. 
\end{itemize}

\subsection{Retrieval Augmented Generation (RAG)}
RAG\cite{RAG} operates by integrating a neural retriever with a sequence-to-sequence (seq2seq) generator. The retriever produces a distribution, $p_{r}(z|x)$ from a dense vector index the fine-tuning dataset based on the input query. These documents then serve as additional context for the seq2seq generator, enabling it to produce outputs that are informed by the retrieved information, $z$, in addition to the user's input, $x$. RAG's seq2seq probability distribution is defined as: 

\begin{align}
p(y|x) &\approx \sum_{z \in \text{top-}k(p_r(\cdot|x))} p_{r}(z|x) p_{Llama}(y|x, z)
\end{align}

This method combines the strengths of pre-trained parametric models with non-parametric external knowledge sources. For our work, we use the pre-trained MiniLM model \cite{miniLM} as the retriever and the pre-trained Llama-2-7B as the generator.

\section{Fine-tuning Dataset and the RAQ Benchmark}

Our fine-tuning dataset consists of several well-known textbooks on microelectronics, VLSI circuit design, and fabrication technologies. In addition to these, we also included some recently published works related to DDR5 design and its corresponding JEDEC standard. After filtering the data, the number of tokens was calculated using Llama-2 tokenizer. The dataset contains 3,168,414 tokens and 12,988 unique tokens. Due to copyright reasons, we cannot release the fine-tuning dataset. However, we list all the components of the dataset within the appendix so that readers can assemble it themselves. 

We created a benchmark to evaluate the performance of the different models which includes 70 carefully-curated domain-specific questions. Among them, there are 40 qualitative questions and 25 true/false questions. The 65 aforementioned questions are meant to evaluate the \textit{accuracy} and \textit{quality} of the LLM's responses on domain knowledge. Furthermore, 5 questions are designed to evaluate the models' capabilities to \textit{reason} upon circuit design decisions based on given specifications. Hence, we name it the \textit{Reasoning-Accuracy-Quality} (RAQ) benchmark. 

\section{Setup and Experiments} 

To assess the suitability of various adaptation methods, we performed four experiments using the RAQ Benchmark:

\begin{enumerate}
    \item \textbf{Student Survey.} We selected 15 questions which would be most relevant for a third-year undergraduate microelectronics classroom. We recorded the responses from each configuration and asked students to provide the ordinal rankings in terms of what they preferred. To ensure quality, we kept the configurations anonymous and asked students to explain why they ranked the best/worst models as they did. After pruning low-quality submissions, we had 51 rankings.
    \item \textbf{True/False Q\&A.} We prompted each configuration to answer true or false to determine accuracy. This portion was taken from the T/F section.
    \item \textbf{Likert Test.} Each configuration was asked to answer all 40 qualitative questions. Using a 7-point Likert scale, the responses were scrutinized in terms of \textit{accuracy} and subjective \textit{quality}. We\footnote{We recognize there could be bias if we, the authors, evaluate these models. To promote transparency, we release the model responses on our GitHub} (human expert) and ChatGPT-4.5 Turbo were the evaluators. 
    \item \textbf{Reasoning Test.} We tested each configuration with 5 reasoning questions. These questions have unambiguous or numerical answers. Generated responses were compared against ground truth values. 
\end{enumerate}

For all experiments, we use $\mathtt{LoRA\_Rank = 4}$, the $\mathtt{Adam(\eta=10^{-4})}$ optimizer \cite{adamOpt}, a sequence length of 128 and a batch size of 16. All the models underwent fine-tuning with LoRA for a total of 20 epochs. Regarding the selection of checkpoints for the models: the 16th epoch checkpoint was chosen for the 7B LoRA model, the 17th epoch checkpoint was utilized for the 70B LoRA (teacher) model, and the 14th epoch checkpoint was selected for the 7B LoRA-KD (student) model. These checkpoints were all selected via early stopping. We set $\alpha = 0.80$ and temperature = $2.0$ for KD.

\section{Results and Conclusion}

In this work, we try to explore the feasibility of using language models in EDA education. Table \ref{table:reasoning_accuracy} investigates the configurations' capabilities to reason based on the given information and optimize a given design. A few interesting observations were made while evaluating the models on reasoning/optimization questions.
\begin{enumerate}
    
    \item All the models had difficulty with numerical calculations and assigning proper units to a calculated value.
    \item In our experiments, the models performed better when they were asked the different sections of the questions one by one in separate prompts, rather than putting all the questions in a single prompt.
    \item RAG tended to refuse answering due to dubious ethical reasons regarding the ``danger of transistors.'' 
\end{enumerate}

An analysis of the data presented in tables \ref{table:reasoning_accuracy} and \ref{tab:likert} gives insights into the strengths and weaknesses of the configurations. For the Likert test and true/false accuracy, 7B RAG performs strongly but for reasoning/optimization, it exhibits a sharp decline in performance. This indicates that RAG alone cannot improve performance across all areas. On the contrary, the various fine-tuned versions manage to perform well on the reasoning portion while remaining within half a standard deviation from RAG on the Likert test. 

The responses collected from the students underscore each configuration's communication skills and human expectations which can serve as an important guideline when fine-tuning an LLM. An improvement of LoRA-KD over LoRA can be observed in figure \ref{fig:survey} where the responses generated by 7B LoRA-KD were far less likely to be ranked last. Another interesting facet is the agreement between the students with respect to the question (i.e. the entropy). For example, for Q14, there was high agreement that the 70B Baseline did the worst. On the other hand, for Q15, the students seemed split between whether 7B RAG, 7B LoRA, or 70B LoRA was the worst.

While the existing works are significant milestones of the application of LLMs in EDA, its potential in this field has yet to be fully realized. Development of specialized large language models capable of understanding the intricacies of domain-specific EDA tasks is crucial for its continued applications in EDA \cite{mirage}. This study highlights some strengths and weaknesses of different open-source offline LLM configurations.

\clearpage
\newpage

\bibliographystyle{plain}

\vspace{12pt}

\onecolumn

\section{Appendix}\label{sec: appendix}

\subsection{Fine-tuning Sources}
The following sources were used in fine-tuning. An enumerated list is also available on our github:

\begin{enumerate}
    \item Fundamentals of Microelectronics - 2nd Edition - Behzad Razavi
    \item Electronic Devices and Circuit Theory - 11th Edition - Robert L. BoyleStad and Louis Nashelsky
    \item CMOS VLSI Design - 4th Edition - Neil H. E. Weste and David M. Harris
    \item Fundamentals of Semiconductor Manufacturing and Process Control - Gary S. May and Costas J. Spanos
    \item Fabrication Engineering at the Micro and Nanoscale - 3rd Edition - Stephen A. Campbell
    \item JEDEC Standard - Graphics Double Data Rate (GDDR5) SGRAM Standard
    \item JEDEC Standard - Compression Attached Memory Module (CAMM2) Common Standard
    \item JEDEC Standard - DDR5 Clocked Small Outline Dual Inline Memory Module (CSODIMM) Common Standard
    \item DDR5 Clocked Unbuffered Dual Inline Memory Module (CUDIMM) Common Specification
    \item JEDEC Standard - DDR5 262 Pin SODIMM Connector Performance Standard
    \item JEDEC Standard - DDR5 Unbuffered Dual Inline Memory Module (UDIMM) Common Standard
    \item JEDEC Standard - DDR5 288 Pin U/R/LR DIMM Connector Performance Standard
    \item JEDEC Standard - DDR5 Load Reduced (LRDIMM) and Registered Dual Inline Memory Module (RDIMM) Common Specification
    \item JEDEC Standard - DDR5 Clock Driver Definition (DDR5CKD01)
    \item JEDEC Standard - DDR5 Small Outline Dual Inline Memory Module (SODIMM) Common Standard
    \item JEDEC Standard - DDR5 Registering Clock Driver Definition (DDR5RCD03)
    \item JEDEC Standard - DDR5 DIMM Labels
    \item JEDEC Standard - GDDR5 Measurement Procedures
    \item JEDEC Standard - DDR5 Serial Presence Detect (SPD) Contents
    \item JEDEC Standard - Graphics Double Data Rate (GDDR5X) SGRAM Standard
    \item JEDEC Standard - DDR5 SDRAM
    \item Improving Memory Reliability by Bounding DRAM Faults - KJERSTEN CRISS, KULJIT BAINS, RAJAT AGARWAL, TANJ BENNETT, TERRY GRUNZKE, JANGRYUL KEITH KIM, HOEJU CHUNG, MUNSEON JANG
    \item Optimizing DDR5 address signal integrity using stochastic learning algorithms - Nitin Bhagwath, Daniel DeAraujo, Jayaprakash Balachandran, BaekKyu Choi 
    \item DDR5 Electrical Challenges in High-Speed Server Design - Douglas Winterberg, Vijender Kumar, Tom Chen, Bhyrav Mutnury
    \item Modeling of DDR5 Signaling from Jitter Sequences to Accurate Bit Error Rate (BER) - Alaeddin A. Aydiner, Yunhui Chu, Oleg Mikulchenko, Jin Yan, Robert J. Friar, Ellen Yan Fu
    \item LPDDR5 (6.4 Gbps) 1-tap DFE Optimal Weight Determination - Sunil Gupta, Ph.D.
    \item Far-End Crosstalk Mitigation for Transmission Lines in DDR5 Using Glass-Weave Coating Structure - Xiao-Bo Yu, Qiang-Ming Cai, Liang Zhang,
Chao Zhang, Lin Zhu, Xin Cao, and Jun Fan
    \item Simulating DDR5 Systems with Clocked Receivers - Matthew Leslie, Justin Butterfield, Randy Wolff
    \item Design and Analysis of Power Integrity of DDR5 Dual In-Line Memory Modules - Shinyoung Park, Vinod Arjun Huddar
    \item Deterministic Policy Gradient-based Reinforcement Learning for DDR5 Memory Signaling Architecture Optimization considering Signal Integrity - Daehwan Lho, Hyunwook Park, Keunwoo Kim, Seongguk Kim, Boogyo Sim, Kyungjune Son, Keeyoung Son, Jihun Kim, Seonguk Choi, Joonsang Park, Haeyeon Kim, Kyubong Kong, Joungho Kim
    \item Advancing DDR5 Test and Measurements: Fine-tuning a Large Language Model AI Expert in DDR5 Protocols - Xinran Li
    \item DDR5 Design Challenges - Nitin Bhagwath, Randy Wolff, Shinichiro Ikeda, Eiji Fujine, Ryo Shibata, Yumiko Sugaya, Megumi Ono
    \item Advanced Measurement and Simulation Approach for DDR5 On-chip SI/PI with the Probing Package - WonSuk Choi, SangKeun Kwak, Jaeseok Park, Jiyoung Do, Byeongseon Yun, Yoo-jeong Kwon, Dongyeop Kim, Kyudong Lee, Tae young Kim, Wonyoung Kim, Kyoungsun Kim, Sung Joo Park, Jeonghyeon Cho and Hoyoung Song
    
\end{enumerate}

\end{document}